\documentclass[10pt, conference, letterpaper]{IEEEtran}
\IEEEoverridecommandlockouts
\usepackage{cite}
\usepackage{amsmath,amssymb,amsfonts, mathtools, array}
\usepackage{graphicx}
\usepackage{textcomp}
\usepackage{xcolor}
\usepackage{caption}
\usepackage{subcaption}
\usepackage{tabularx}
\usepackage{float}
\usepackage{lipsum,mathtools}
\usepackage{multirow}
\usepackage{dirtytalk}
\usepackage{hyperref}
\usepackage{amssymb, nccmath}
\usepackage[linesnumbered, ruled,vlined]{algorithm2e}
\usepackage[numbers, sort&compress]{natbib}
\def\BibTeX{{\rm B\kern-.05em{\sc i\kern-.025em b}\kern-.08em
    T\kern-.1667em\lower.7ex\hbox{E}\kern-.125emX}}
    \IEEEaftertitletext{\vspace{-2\baselineskip}}
    
\begin{document}
\title{Embedded Federated Feature Selection with Dynamic Sparse Training: Balancing Accuracy-Cost Tradeoffs}
\author{\IEEEauthorblockN{Afsaneh Mahanipour}
\IEEEauthorblockA{\textit{Department of Computer Science} \\
\textit{University of Kentucky}\\
Lexington, KY, USA \\
ama654@uky.edu}
\and
\IEEEauthorblockN{Hana Khamfroush}
\IEEEauthorblockA{\textit{Department of Computer Science} \\
\textit{University of Kentucky}\\
Lexington, KY, USA \\
khamfroush@cs.uky.edu}
}
\maketitle
\begin{abstract}
 Federated Learning (FL) enables multiple resource-constrained edge devices with varying levels of heterogeneity to collaboratively train a global model. However, devices with limited capacity can create bottlenecks and slow down model convergence. One effective approach to addressing this issue is to use an efficient feature selection method, which reduces overall resource demands by minimizing communication and computation costs, thereby mitigating the impact of struggling nodes. Existing federated feature selection (FFS) methods are either considered as a separate step from FL or rely on a third party. These approaches increase computation and communication overhead, making them impractical for real-world high-dimensional datasets. To address this, we present \textit{Dynamic Sparse Federated Feature Selection} (DSFFS), the first innovative embedded FFS that is efficient in both communication and computation. In the proposed method, feature selection occurs simultaneously with model training. During training, input-layer neurons, their connections, and hidden-layer connections are dynamically pruned and regrown, eliminating uninformative features. This process enhances computational efficiency on devices, improves network communication efficiency, and boosts global model performance. Several experiments are conducted on nine real-world datasets of varying dimensionality from diverse domains, including biology, image, speech, and text. The results under a realistic non-iid data distribution setting show that our approach achieves a better trade-off between accuracy, computation, and communication costs by selecting more informative features compared to other state-of-the-art FFS methods.
\end{abstract}

\begin{IEEEkeywords}
Dynamic Sparse training, Feature selection, Federated learning
\end{IEEEkeywords}

\section{Introduction}
Federated Learning (FL) is a decentralized machine learning technique that allows edge devices/clients with limited resources and different levels of heterogeneity to jointly train a global model. During this process, only model parameters are iteratively exchanged through a central cloud server, ensuring that local data remains private \cite{kairouz2021advances}. FL is categorized into two main scenarios: horizontal FL and vertical FL \cite{zhu2021federated}. In vertical FL, clients have datasets that contain the same instances but with different feature sets \cite{li2023fedsdg, castiglia2023less}. In contrast, horizontal FL involves clients with different instances that share the same feature set \cite{huang2022fairness}. Horizontal FL is extensively applied in real-world scenarios. For instance, self-driving cars can collaboratively improve obstacle detection and navigation systems by sharing learned insights without revealing raw sensor data. Similarly, hospitals can enhance diagnostic models by exchanging knowledge derived from patient data while maintaining data privacy \cite{li2020review}.

Clients often generate or collect vast amounts of high-dimensional data, which may contain noisy, irrelevant, or redundant features. These features can add computational overhead, increase memory usage, extend execution time, and raise communication costs between clients and the server, ultimately degrading model performance. In the FL process, resource-limited clients can become bottlenecks, slowing down model convergence. An effective way to address this challenge is through efficient feature selection, which reduces resource demands by identifying and retaining the most informative features. By decreasing data size, it minimizes communication and computation costs, alleviates the impact of struggling nodes, and enhances overall system efficiency \cite{mahanipour2023multimodal}.

Most feature selection (FS) methods are designed for centralized settings, with only a few addressing federated FS in horizontal FL. Centralized FS methods are unsuitable for horizontal FL as they either require direct access to data or are inefficient in terms of computation and communication. Existing federated feature selection (FFS) methods for horizontal FL often function as a separate step from FL or rely on a third party, leading to increased computational and communication costs, making them impractical for real-world high-dimensional datasets \cite{cherepanova2024performance}.

FS methods are generally classified into three categories: filter methods, wrapper methods, and embedded methods. Existing FFS approaches rely on filter- or wrapper-based techniques, which are less effective in selecting informative features compared to embedded methods. Filter-based methods evaluate and rank features using inherent data characteristics (e.g., information theory) \cite{cassara2022federated}, while wrapper methods assess feature subsets using learning models, resulting in high computational costs \cite{hu2022multi, mahanipour2023wrapper}. To overcome these limitations, this work aims to answer the following research question: \textit{Can we design an embedded-based, computation- and communication-efficient federated feature selection method for horizontal FL that selects informative features while achieving a good trade-off between accuracy, computation, and communication costs?}



\begin{figure}
\includegraphics[width=\linewidth]{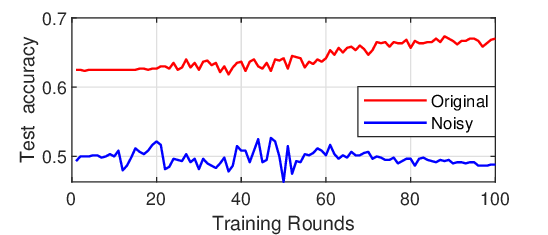}
\centering
\caption{Test accuracy of FedDST on an artificial dataset with original and noisy features.}
\label{fig1}
\end{figure}

Classical horizontal FL methods distribute dense model parameters across clients for local training. After training, these local parameters are sent to a central server for aggregation and updating of the global model \cite{luo2021cost}. However, this process leads to high communication and computation costs, along with substantial memory demands due to the large parameter sizes. To address these challenges, significant efforts have been made to optimize lightweight neural networks in FL. One such approach is Federated Dynamic Sparse Training (FedDST) \cite{bibikar2022federated}, which trains a sparse subset of the model on each client, significantly reducing computation and communication costs. However, the presence of non-informative features can degrade the global model's performance in FedDST, increasing communication overhead, computational costs, and memory consumption. Figure \ref{fig1} illustrates this issue by comparing the test accuracy of FedDST on an original artificial dataset versus a noisy version, demonstrating a substantial performance drop in the presence of noise.

In this work, we introduce DSFFS (Dynamic Sparse Federated Feature Selection), the first embedded-based sparse FFS method. DSFFS incorporates a dynamic updating mechanism for input-layer neurons and connections, enhancing the performance of the FL algorithm. By integrating feature selection directly into the training process while maintaining minimal communication and computation costs, DSFFS is highly efficient for resource-constrained clients. It leverages the weights of sparse connections to identify the most relevant features efficiently. Our key contributions are as follows:

\begin{itemize}
\item Introducing the first embedded-based FFS method for FL.
\item Integrating a dynamic pruning and regrowth mechanism in the input layer of a sparse FL to enable efficient FS.
\item Conducting extensive experiments on nine real-world datasets from diverse domains, demonstrating that our approach provides a superior trade-off between model accuracy, communication, and computation costs compared to state-of-the-art FFS methods.
\item Improving computational and communication efficiency while preserving accuracy by leveraging sparse models.
\end{itemize}

\section{RELATED WORKS}
Previous works have mainly concentrated on centralized feature selection methods, with only a few studies exploring federated feature selection in horizontal FL.

\subsection{Centralized Feature Selection}
Feature selection methods are generally categorized into three main types: filter, wrapper, and embedded methods. Filter methods are considered as a pre-processing step, ranking features based on inherent data characteristics such as information theory, correlation and Gini index before training the learning model. These methods are fast but often select redundant features \cite{zebari2020comprehensive}. Wrapper methods use various search strategies, like evolutionary algorithms, incorporating feedback from learning models to evaluate the quality of selected feature sets. Therefore, these methods have higher computational costs and are not suitable for high-dimensional datasets. Embedded methods integrate the feature selection process with the learning model itself, using approaches like support vector machine classifiers and neural networks to select features. These methods are more accurate than filter methods and require less computational time and complexity compared to wrapper methods \cite{dokeroglu2022comprehensive}. 
Centralized FS methods assume all data is stored in a single location, making them unsuitable for distributed scenarios like horizontal FL, where privacy concerns, communication, and computation costs are critical.


\subsection{Federated Feature Selection in Horizontal FL}
There are only a few studies available in the literature on supervised federated feature selection in horizontal FL. In \cite{cassara2022federated}, a novel filter-based FFS method called CE-FFS, which relies on Cross-Entropy, is proposed. This method selects only a small number of essential features without considering the learning model, leading to information loss and an inadequate trade-off between communication cost and learning model performance. Additionally, GSA-FFS \cite{mahanipour2023wrapper} is another wrapper-based method that employs a binary gravitational search algorithm for FS on both the client side and the edge server side. The results indicate that this method can achieve a reasonable trade-off between communication cost and learning model performance. However, the computational cost is high due to the use of an evolutionary algorithm-based FS on the client side. Similarly, Hu et al. \cite{hu2022multi} propose another wrapper FFS algorithm based on particle swarm optimization (PSO). A binary bare-bones PSO is used as a local feature selector to choose features from imbalanced local datasets, which are then assembled by a trusted third party. However, this approach may result in inaccurate FFS since there is no global feature selector on the edge server to continue the evolutionary process. Additionally, it causes high communication costs due to data transmission between clients and the trusted third party. To address these limitations, the proposed method introduces the first embedded FFS approach, offering a more efficient solution for FFS.
\vspace{-2mm}
\subsection{Dynamic Sparse Training}
Dynamic Sparse Training (DST) \cite{mocanu2018scalable} is a method for training sparse neural networks. In DST, a neural network with a random sparse topology, denoted as $f(x,\theta_s)$, is initialized from scratch. Here, \(\theta_s\) represents the parameters of a sparse sub-network, where the sparsity level \(s=1-\frac{\Vert\theta_s\Vert_0}{\Vert\theta\Vert_0}\). \(\Vert \theta_s\Vert_0\) and \(\Vert \theta\Vert_0\) indicate the number of parameters in the sparse and dense networks, respectively. \(\Vert.\Vert_0\) denotes the zero-norm, which counts the non-zero elements in \(\theta\). During training, the sparse topology is periodically updated by removing and adding an equal fraction of parameters to maintain a constant sparsity level \(s\). The goal of DST methods is to optimize the following problem:
\vspace{-1mm}
\begin{equation}\label{my_forth_eqn}
\theta_s^* =\underset{\theta_s \in \mathbb{R}^{\Vert\theta\Vert_0},\, \Vert\theta_s\Vert_0=h\Vert \theta\Vert_0 } {\mathrm{arg\,min}}\, \frac{1}{N}\sum_{i=1}^N J(f(x_i;\theta_s),y_i)
\end{equation}

\noindent where \(h\) represents the density level and is defined as \(h=1-s\), and \(J\) is a desired loss function. In the literature, DST methods have been applied in various fields, including deep reinforcement learning \cite{sokar2021dynamic}, ensembling \cite{liu2021deep}, feature selection \cite{sokar2022pay}, and federated learning \cite{bibikar2022federated}.

\begin{figure*}
\includegraphics[width=0.85\textwidth]{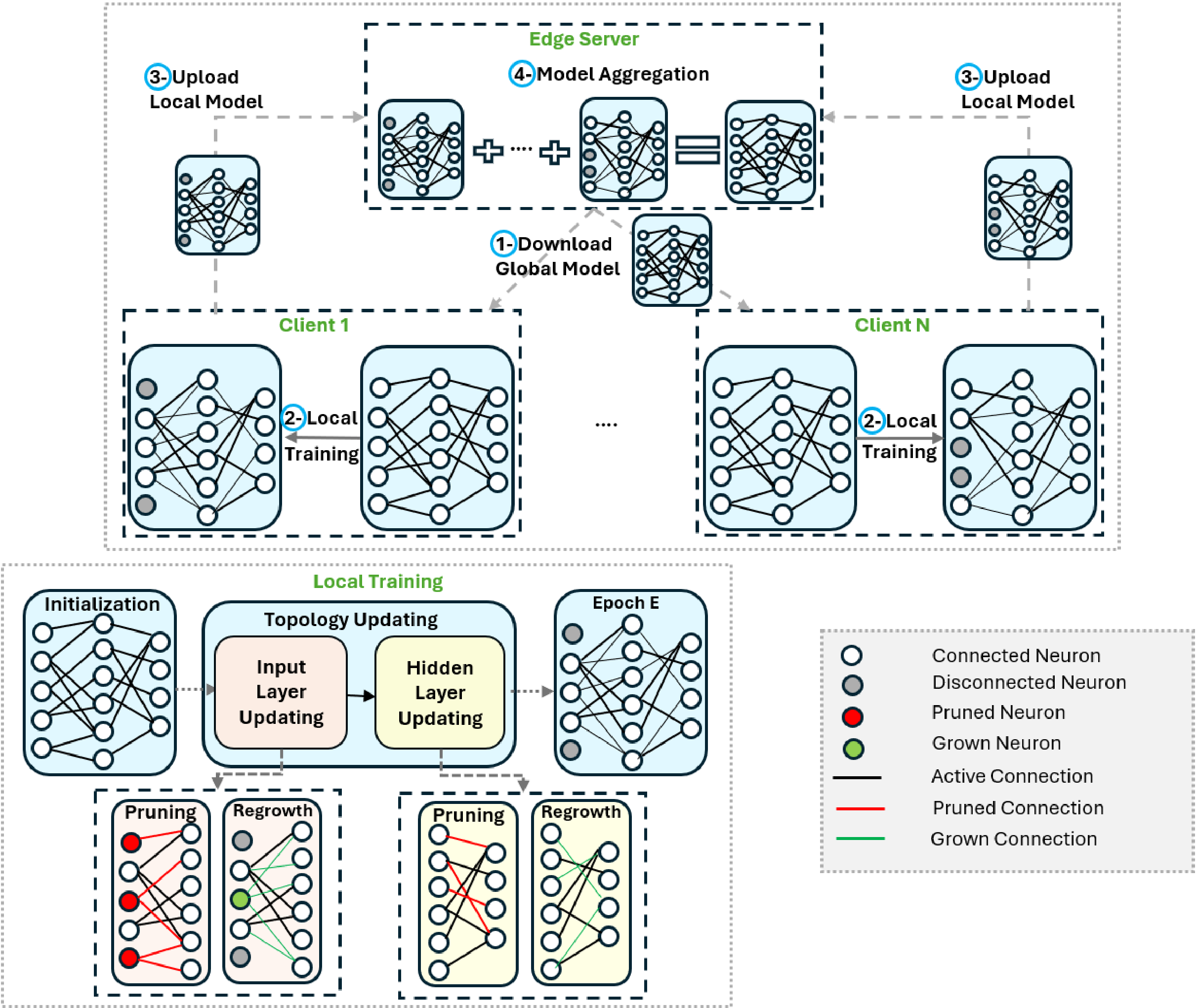}
\centering
\caption{Overview of the proposed method DSFFS for embedded-based federated feature selection.}
\label{fig2}
\end{figure*}

\begin{algorithm}
\caption{Overview of the proposed Dynamic Sparse Federated Feature Selection (DSFFS)}\label{alg:cap}
\KwIn{$M$ Clients with their local datasets $\mathcal{U}_i$, Sparsity level $s$, \(\beta\), \(\zeta\), Number of local training epochs $Q$, Number of rounds $r_{max}$, Desired number of selected features $K$}
{\textbf{Initialization:} Initialize the edge server model with sparsity level ($s$), \(\Vert{\theta_s^1}\Vert_0\)}\;
\For{each round $r$}{
    Transmit the server sparse network as the global network to all clients (\(M\))\;
    \For{each client \(m \in M\)} {
        Receive the global model from the server (\(\theta_{s_m}^r \leftarrow \theta_s^r\))\;
        \For{each local training epoch \(q \in Q\)}{%
            Select a minibatch from local dataset \(\mathcal{U}_m\)\;
            Perform training of local sparse network\;
            \Indp
            \textbf{Update Input Layer:}\;
                \Indp
                Calculate \(n_p^r\) (Eq. 4) and \(n_g^r\) (Eq. 8)\;
                Prune \(n_p^r\) input neurons with the lowest strength\;
                Prune a fraction of input connections with the lowest magnitude\;
                Reconnect \(n_g^r\) disconnected input neurons with connections having the highest gradient magnitudes\;
                Re-add the same number of connections that were pruned\;
                \Indm
            \textbf{Update Hidden Layers:}\; 
                \Indp
                Prune a fraction of connections using layer-wise magnitude pruning\;
                Regrow an equal number of connections using layer-wise gradient magnitude growth\;
                \Indm
            \Indm
        \textbf{end for}
        }
        Transmit the new \(\theta_m^{\prime r}\) to the server\;
    \textbf{end for}
    }
    Receive all local sparse networks from clients \(m \in M\)\;
    Aggregate all networks (Eq. 2)\;
\textbf{end for}
}
\end{algorithm}

\section{Proposed Method}
\subsection{System Setup and Problem Formulation}
In this setup, we design a two-tier system for our DSFFS method. The first tier comprises \(M\) clients, represented as \(C_m\) (where \(m=1,2,...,M\)). The second tier includes an edge server, denoted as \(e\), situated closer to the clients. While the proposed approach is implemented with an edge server, it can be easily generalized to scenarios where the edge server is replaced by a central cloud. Importantly, \(M\) must be at least 2, as having only a single client would reduce the problem to a centralized FS scenario. Each client's dataset is represented as \(\mathcal U_{m}=\{X, Y\}=\{(x_i, y_i)\}_{i=1}^{N_{m}}\), where \(N_{m}\) is the number of instances for client $m$, and these values may differ across clients. Each instance consists of a \(D\)-dimensional feature vector \(x_i=(x_{i1}, x_{i2}, ..., x_{iD})\) and its corresponding label \(y_i\).

In federated DST, the edge server initializes a random sparse network \(\theta_s^1\) at round \(r=1\) and applies the layer-wise magnitude pruning and gradient-magnitude weight growth techniques \cite{evci2020rigging}. In each round, the server selects a group of clients and sends the sparse network to them. Each client then trains a local model for \(Q\) epochs on their local data \(\mathcal U_m\) and performs a network readjustment procedure to identify a more effective sub-network. After local training, clients send their updated sparse networks back to the server, which aggregates them using a sparse weighted average:

\begin{equation}\label{my_forth_eqn}
\theta_s^{r+1}=\frac{\sum_{m=1}^{M}N_m\theta_{s_m}^r}{\sum_{m=1}^{M}N_m}
\end{equation}

We propose DSFFS, a novel embedded-based supervised federated feature selection method utilizing federated DST. The objective is to collaboratively select a subset of \(K\) informative features across all clients while optimizing the sparse topology of the global network. DSFFS updates the input layer neurons and connections during training by dynamically dropping and adding neurons based on their strength, thereby selecting the \(K\) input neurons with the highest strength. An overview of DSFFS is demonstrated in Figure \ref{fig2}. The density of each layer \(h_l\) is defined as \(h_l=\frac{s(n^{l-1}+n^l)}{(n^{l-1}\times n^l)}\), where \(l \in \{1, 2, ..., L\}\) represents the layer index, \(s\) is the sparsity level, and \(n^l\) is the number of neurons in layer \(l\). The total number of parameters in the sparse network is given by \(\Vert{\theta_s}\Vert_0 = \sum_{l=l}^L \Vert{\theta_s^l}\Vert_0\), where \(\Vert{\theta_s^l}\Vert_0 = s(n^{l-1}+n^l)\). During the training process, the sparsity level remains fixed, and the input layer along with the sparse topology is optimized to select the most discriminative features. The following section details the training algorithm for each local sparse network.

\subsection{Training Algorithm}
In DSFFS, the input layer and hidden layers are dynamically updated to improve the performance of local networks within federated DST, enabling efficient and effective selection of informative features. The input layer update process occurs in two steps during each epoch: pruning and regrowth. In the pruning step, input neurons with the weakest connections and the lowest magnitudes are removed from the network. Following this, the regrowth step reintroduces a specified number of previously unconnected neurons back into the network.

\textbf{Input Layer Pruning Step:} At each epoch, the strength of all input neurons is calculated using Eq. \ref{my_3_eqn}, which is inspired by graph theory \cite{atashgahi2022quick, atashgahi2023supervised}. The strength of a neuron is defined as the sum of the absolute weights of its connections. A higher strength value indicates that the neuron corresponds to a more informative and distinctive feature. Subsequently, a number of input neurons \(n_p\) with the lowest strength values are pruned. If \(w_i\) represents the set of weights associated with neuron \(i\) in the input layer, the strength of this neuron is calculated as follows:

\begin{equation}\label{my_3_eqn}
H_i=\Vert{w_i}\Vert_1
\end{equation}

\noindent Next, a fraction \(\zeta\) of the input layer connections from the remaining connected input neurons are pruned (Lines 9-12 of Algorithm 1).

\textbf{Input Layer Regrowth Step:} After pruning a subset of input neurons, the disconnected neurons are evaluated to identify the most informative ones for reconnection. This evaluation is based on the gradient magnitudes of their connections, with the neurons having the highest absolute gradients being reconnected. This gradient-magnitude-based growth operation is a lightweight process that accelerates learning and improves overall performance.

After regrowing \(n_g\) disconnected input neurons based on the highest absolute gradients, the next step is to regrow input connections. To maintain a consistent sparsity level during training, the number of new connections must equal the number of pruned connections. These new connections are selected by identifying the highest absolute gradients among all currently non-existent connections of the connected neurons.

The number of input neurons to be pruned \((n_p)\) and regrown \((n_g)\) in each round \(r\) are determined using the following equations \cite{atashgahi2023supervised}:

\begin{equation}\label{my_forth_eqn}
n_p^r = 
    \begin{cases}
    n_{remove}^r + n_g^r, & {r \leq r_{remove}}\\
    n_g^r, & \text{otherwise}
    \end{cases}
\end{equation}

\noindent where \(r_{remove}=\lceil \beta \times r_{max} \rceil\), \(0<\beta<1\), and \(r_{max}\) represents the maximum number of rounds. In this method, from the first round up to \(r_{remove}\), more input neurons are pruned than regrown in the input layer. Beyond \(r_{remove}\), from \((r_{remove}+1)\) to \(r_{max}\), the number of neurons pruned equals the number of neurons regrown.

Here, \(n_{remove}^r\) and \(n_g^r\) are calculated as follows:

\begin{equation}\label{my_forth_eqn}
n_{remove}^r = \lceil \frac{T - T^r}{r_{remove} - r} \rceil,
\end{equation}

\begin{equation}\label{my_forth_eqn}
T^r = \sum_{i=1}^{r-1} n_{remove}^i,
\end{equation}

\begin{equation}\label{my_forth_eqn}
T = \lceil (1-\zeta)D-K \rceil,
\end{equation}

\begin{equation}\label{my_forth_eqn}
n_g^r = \lceil \zeta(1-\frac{r}{r_{max}}) T^r \rceil.
\end{equation}

\noindent where \(T^r\) represents the total number of disconnected input neurons in round \(r\), while \(T\) denotes the number of input neurons pruned in each round. During the regrowth step, a linearly decreasing fraction of these disconnected neurons is reconnected as the training progresses, ensuring the network is continuously updated (Lines 13, 14 in Algorithm 1).

After updating the input layer, the hidden layers should be updated. This process begins with layer-wise magnitude pruning, where connections with the smallest magnitudes are removed. Subsequently, an equal number of connections with the largest gradients are added to maintain the sparsity level (Lines 15-17 in Algorithm 1).

\textbf{Feature Selection:} After the training process, the \(K\) most distinctive and informative features are selected from the \(\zeta D+K\) connected input neurons based on their strength values. The detailed steps of the proposed method are outlined in Algorithm 1.

\section{Experimental Results}
\vspace{-2mm}
\subsection{Datasets}
The performance of the proposed method is evaluated using 9 publicly available datasets\footnote[1]{Available at \url{https://jundongl.github.io/scikit-feature/datasets.html}} from various domains, including biology, time series, image, speech, and text, each characterized by different numbers of features and instances. Table \ref{table1} provides detailed information about these datasets.

\begin{table}[htbp]
\caption{Details of the benchmark datasets.}
\vspace{-3mm}
\label{table1}
\begin{center}
\resizebox{\columnwidth}{!}{%
\begin{tabular}{c c c c c}
\hline
Domain & Dataset & \#Features & \#Instances & \#Classes \\
\hline
 
\multirow{2}{*}{Biology} & \texttt{SMK-CAN-187} & 19993 & 187 & 2\\
 & \texttt{GLA-BRA-180} & 49151 & 180 & 4\\
\hline
Time Series & \texttt{HAR} & 561 & 10299 & 6\\
\hline
\multirow{4}{*}{Image}  & \texttt{COIL-20} & 1024 & 1440 & 20\\
 & \texttt{USPS} & 256 & 9298 & 10\\
 & \texttt{MNIST} & 784 & 70000 & 10\\
 & \texttt{Fashion-MNIST} & 784 & 70000 & 10\\
\hline
Speech & \texttt{Isolet} & 617 & 7737 & 26\\
\hline
Text & \texttt{PCMAC} & 3289 & 1943 & 2\\
\hline
\end{tabular}}
\end{center}
\end{table}

\begin{table*}[htbp]
\caption{Comparison of the proposed method with three state-of-the-art supervised FFS methods from the literature in terms of accuracy and \#FLOPs (\(10^{9}\)).}
\vspace{-3mm}
\label{table4}
\setlength{\tabcolsep}{0.85\tabcolsep}
\begin{center}
\begin{tabular}{c c c |c c c |c c c |c c c}
\hline
 &  & & \multicolumn{3}{c|}{\texttt{COIL-20}} & \multicolumn{3}{c|}{\texttt{USPS}} & \multicolumn{3}{c}{\texttt{MNIST}} \\
\hline 
Training Method & FFS Method & Type & \#Features & Accuracy & FLOPs & \#Features & Accuracy & FLOPs & \#Features & Accuracy & FLOPs \\ \hline

\multirow{4}{*}{FedDST} & DSFFS & Embedded & 150 & \textbf{0.8298} & 0.1147 &  100 & \textbf{0.8516} & \textbf{0.0865} & 150 & \textbf{0.8051} & \textbf{0.2243}  \\
& MFPSO & Wrapper & 271 & 0.7395 & 0.1766 & 110 & 0.7951 & 0.0916 & 309 & 0.7677 &  0.3871  \\
& GSA-FFS & Wrapper & 505 & 0.7534 & 0.2964 & 123 & 0.7849 & 0.0983 &  403 & 0.7825 & 0.4833 \\
& CE-FFS & Filter & 123 & 0.6770 & \textbf{0.1009} & 129 & 0.7913 & 0.1014 & 386 & 0.7633 &  0.4659 \\
\hline
\multirow{4}{*}{FedDST+FedProx} & DSFFS & Embedded & 150 & \textbf{0.8090} & 0.1147 &  100 & 0.8112 & \textbf{0.0865} & 150 & \textbf{0.7933} & \textbf{0.2243}  \\
& MFPSO & Wrapper & 271 & 0.75 & 0.1766 & 110 & \textbf{0.8295} & 0.0916 & 309 & 0.7702 &  0.3871  \\
& GSA-FFS & Wrapper & 505 & 0.7569 & 0.2964 & 123 & 0.7715 & 0.0983 &  403 & 0.7807 & 0.4833 \\
& CE-FFS & Filter & 123 & 0.6701 & \textbf{0.1009} & 129 & 0.7956 & 0.1014 & 386 & 0.7641 &  0.4659 \\
\hline
 &  & & \multicolumn{3}{c|}{\texttt{Fashion-MNIST}} & \multicolumn{3}{c|}{\texttt{Isolet}} & \multicolumn{3}{c}{\texttt{PCMAC}} \\
\hline
\multirow{4}{*}{FedDST} & DSFFS & Embedded & 150 & \textbf{0.7383} & \textbf{0.2243} &  150 & \textbf{0.6891} & \textbf{0.1162} & 150 & \textbf{0.8688} &  \textbf{0.2202} \\
& MFPSO & Wrapper & 405 & 0.6895 & 0.4854 & 163 & 0.6474 & 0.1229 & 1163 & 0.7480 &  1.2575  \\
& GSA-FFS & Wrapper & 405 & 0.7059 & 0.4854 & 328 & 0.6602 & 0.2074 &  1663 & 0.7043 & 1.7695 \\
& CE-FFS & Filter & 384 & 0.7074 & 0.4639 & 313 & 0.6410 & 0.1997 & 223 & 0.5578 &  0.2949 \\
\hline
\multirow{4}{*}{FedDST+FedProx} & DSFFS & Embedded & 150 & \textbf{0.7292} & \textbf{0.2243} &  150 & \textbf{0.6314} & \textbf{0.1162} & 150 & \textbf{0.8688} & \textbf{0.2202}  \\
& MFPSO & Wrapper & 405 & 0.7157 & 0.4854 & 163 & 0.5993 & 0.1229 & 1163 & 0.7403 &  1.2575  \\
& GSA-FFS & Wrapper & 405 & 0.7070 & 0.4854 & 328 & 0.625 & 0.2074 & 1663  & 0.7223 & 1.7695 \\
& CE-FFS & Filter & 384 & 0.7117 & 0.4639 & 313 & 0.6025 & 0.1997 & 223 & 0.6323 &  0.2949 \\
\hline

 &  & & \multicolumn{3}{c|}{\texttt{GLA-BRA-180}} & \multicolumn{3}{c|}{\texttt{HAR}} & \multicolumn{3}{c}{\texttt{SMK-CAN-187}} \\
\hline
\multirow{4}{*}{FedDST} & DSFFS & Embedded & 150 & \textbf{0.5} & \textbf{0.1106} &  150 & \textbf{0.9124} & 0.2222 & 150 & \textbf{0.8156} & \textbf{0.1101}  \\
& MFPSO & Wrapper & 10413 & 0.4722 & 5.3652 & 148 & 0.8581 & \textbf{0.2202} & 6899 & 0.4736 &  3.5656  \\
& GSA-FFS & Wrapper & 24484 & 0.1388 & 12.5696 & 278 & 0.8629 & 0.3533 & 9961  & 0.5263 & 5.1333 \\
& CE-FFS & Filter & 3260 & 0.4722 & 1.7029 & 152 & 0.8568 & 0.2243 & 1153 & 0.5263 & 0.6236  \\
\hline
\multirow{4}{*}{FedDST+FedProx} & DSFFS & Embedded & 150 & 0.4722 & \textbf{0.1106} &  150 & \textbf{0.8839} & 0.2222 & 150 & \textbf{0.7631} &  \textbf{0.1101} \\
& MFPSO & Wrapper & 10413 & 0.4444 & 5.3652 & 148 & 0.8238 & \textbf{0.2202} & 6899 & 0.4736 &  3.5656  \\
& GSA-FFS & Wrapper & 24484 & 0.1388 & 12.5696 & 278 & 0.8788 & 0.3533  & 9961  & 0.4736 & 5.1333 \\
& CE-FFS & Filter & 3260 & 0.4722 & 1.7029 & 152 & 0.8445 & 0.2243 & 1153 & 0.5263 &  0.6236 \\
\hline
\end{tabular}
\end{center}
\end{table*}

\begin{figure}[htbp]
  \centering
    \includegraphics[width=\linewidth]{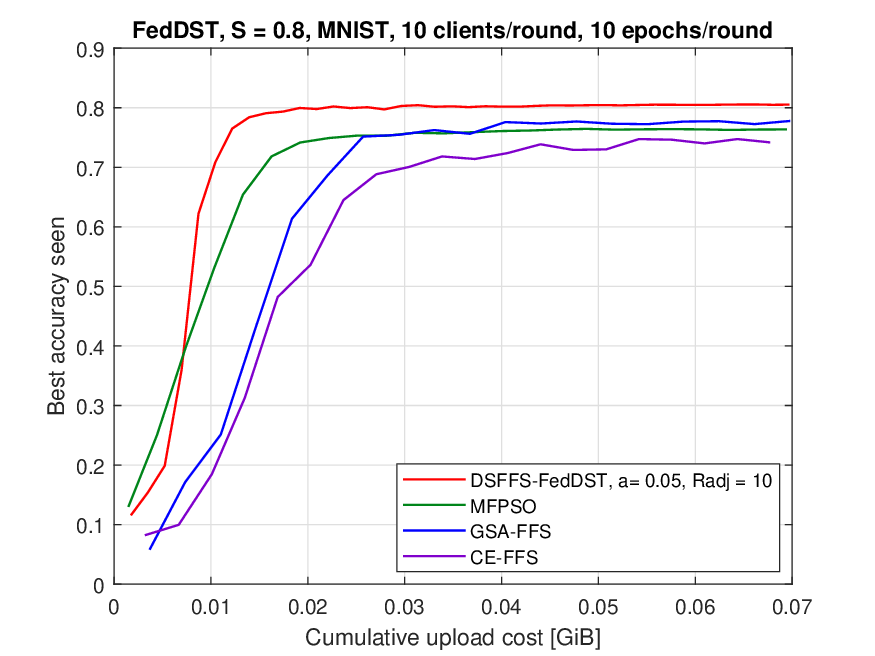}
\caption{Test accuracy vs. Cumulative upload cost on non-iid \(\texttt{MNIST}\).}
\label{fig3}
\end{figure}

\subsection{Baselines}
As previously mentioned, there are only a few FFS methods in horizontal FL, and the proposed method represents the first embedded approach. For comparison, we select three existing supervised FFS methods: \textbf{1- Filter method:} CE-FFS, which uses Cross-Entropy to select informative features  \cite{cassara2022federated}. \textbf{2- Wrapper methods:} GSA-FFS, which employs a binary gravitational search algorithm  \cite{mahanipour2023wrapper}, and MFPSO, which utilizes a trusted third party and a binary bare-bones PSO algorithm on clients to identify important features \cite{hu2022multi}. Additionally, the proposed DSFFS method is integrated into two different FL frameworks, FedDST and FedDST+FedProx \cite{bibikar2022federated}, for performance evaluation. FedDST \cite{bibikar2022federated} is a federated dynamic sparse training approach designed to produce a single model each round that performs effectively across all clients. FedDST+FedProx \cite{bibikar2022federated} builds upon FedDST by incorporating a proximal term and refining the growth criterion adjustments to further enhance performance.

\begin{figure}[htbp]
  \centering
      \includegraphics[width=\linewidth]{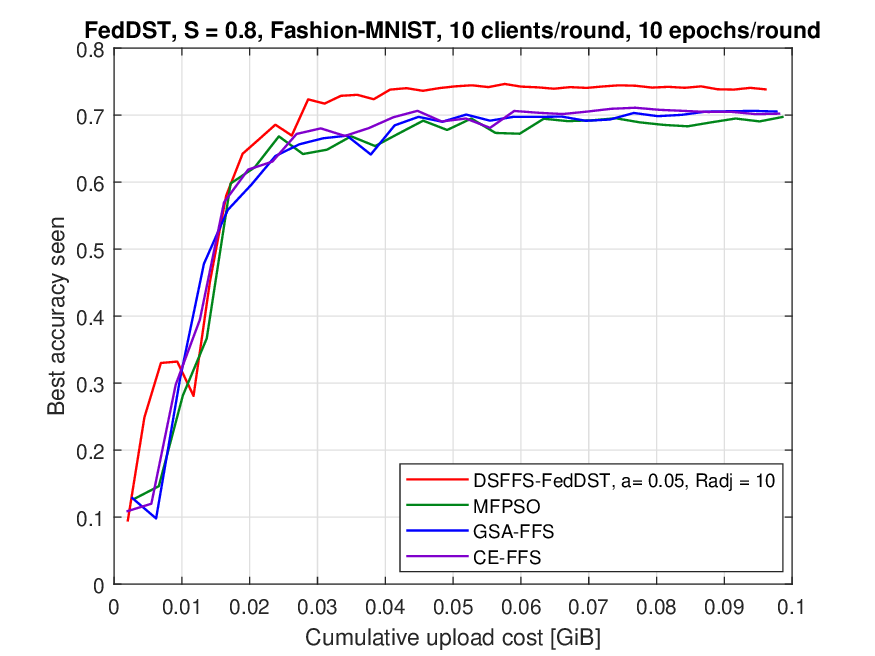}
  \caption{Test accuracy vs. Cumulative upload cost on non-iid \(\texttt{Fashion-MNIST}\).}
  \label{fig4}
\end{figure}

\begin{figure}[htbp]
  \centering
      \includegraphics[width=\linewidth]{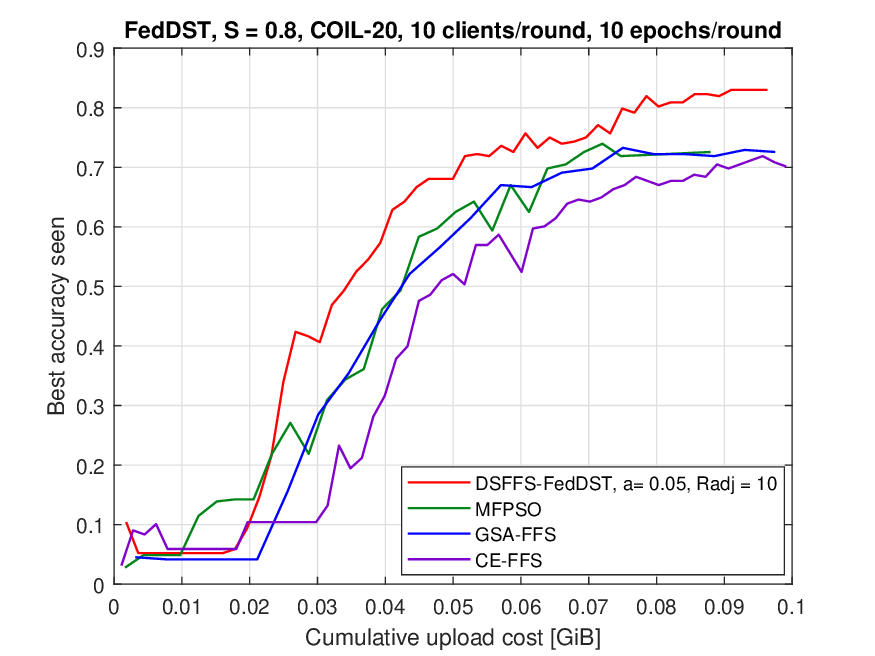}
  \caption{Test accuracy vs. Cumulative upload cost on non-iid \(\texttt{COIL-20}\).}
  \label{fig5}
\end{figure}

\subsection{Evaluation Metrics}
In FFS methods, achieving a balance between model performance, communication cost, and computation cost is crucial. To evaluate the proposed method, three metrics are utilized: \textbf{(1) Classification Accuracy:} To evaluate the informativeness of the features selected by each method, we train an FL model using the selected features by the proposed method and comparative approaches, and report the resulting classification accuracy. \textbf{(2) Computation Cost:} The computational cost of training an FL model on a given dataset is evaluated by calculating the cumulative number of floating-point operations (FLOPs) performed by each client throughout the training process. The FLOPs are computed using the layer-by-layer method detailed in \cite{evci2020rigging}, which accounts for the network's sparsity level. \textbf{(3) Communication Cost:} The cumulative upload cost is calculated based on the number of network parameters. For example, in the FedDST model, the maximum upload or download cost for a client is given by \((32\times(1-S)+1)n\) bits, where \(n\) is the total number of parameters in the network.

\subsection{Parameter setting}
In our experiments, we use 10 clients with non-IID data distributions, consistent with previous FFS methods. The sparsity level is set to 0.8, with \(\beta=0.65\) and \(\zeta=0.2\) for the feature selection process. For the FedDST, we set \(a=0.05\) and \(R_{adj}=10\), while for FedDST+FedProx, we use \(\mu=1\), with all parameters determined through trial and error. Following the methodology outlined in \cite{bibikar2022federated}, we report the classification accuracy after 400 training rounds and 10 local epochs.

\begin{figure*}[htbp]
  \centering
  \begin{tabular}[c]{ccc}
    \begin{subfigure}{0.31\textwidth}
      \includegraphics[width=\textwidth]{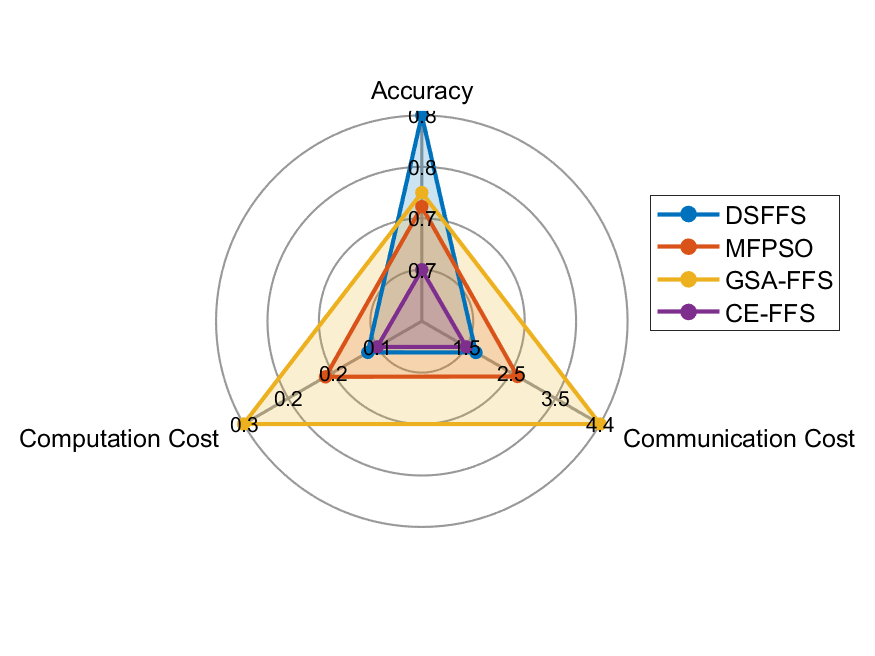}
      \vspace{-13mm}
      \caption{$COIL-20$}
    \end{subfigure}&
    \begin{subfigure}{0.31\textwidth}
      \includegraphics[width=\textwidth]{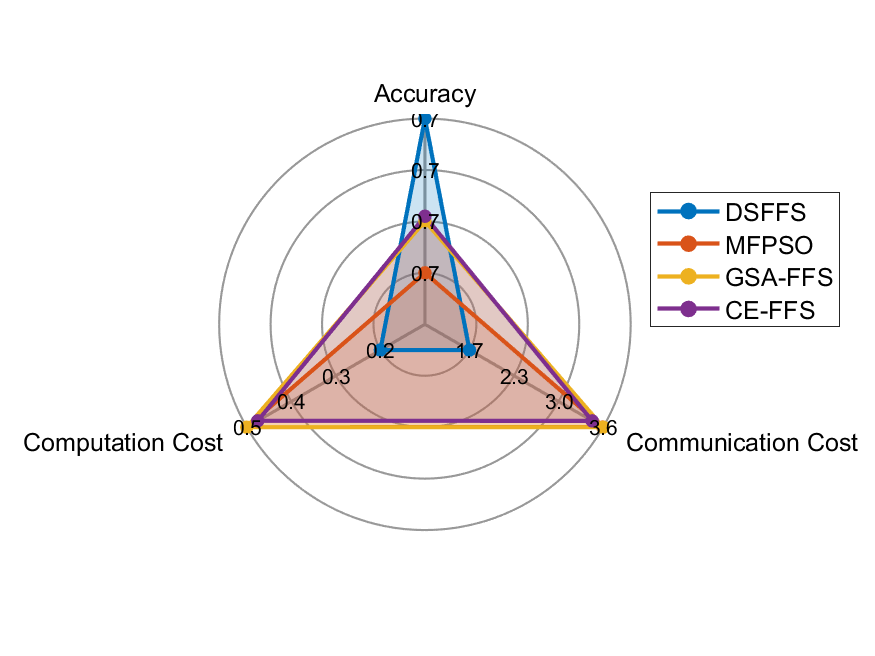}
      \vspace{-13mm}
      \caption{$Fashion-MNIST$}
    \end{subfigure}&
    \begin{subfigure}{0.31\textwidth}
      \includegraphics[width=\textwidth]{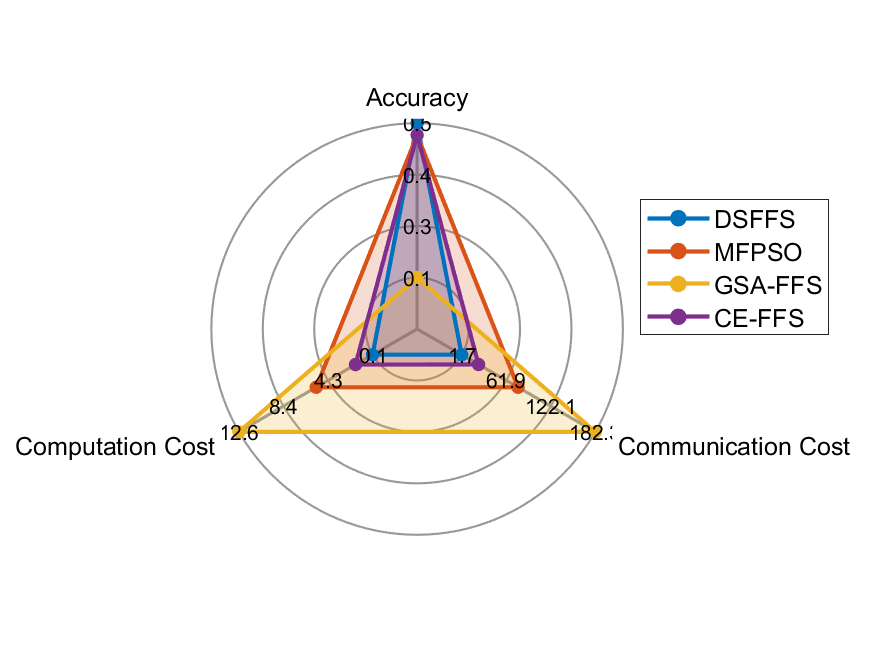}
      \vspace{-13mm}
      \caption{$GLA-BRA-180$}
    \end{subfigure}\\
    \begin{subfigure}{0.31\textwidth}
      \includegraphics[width=\textwidth]{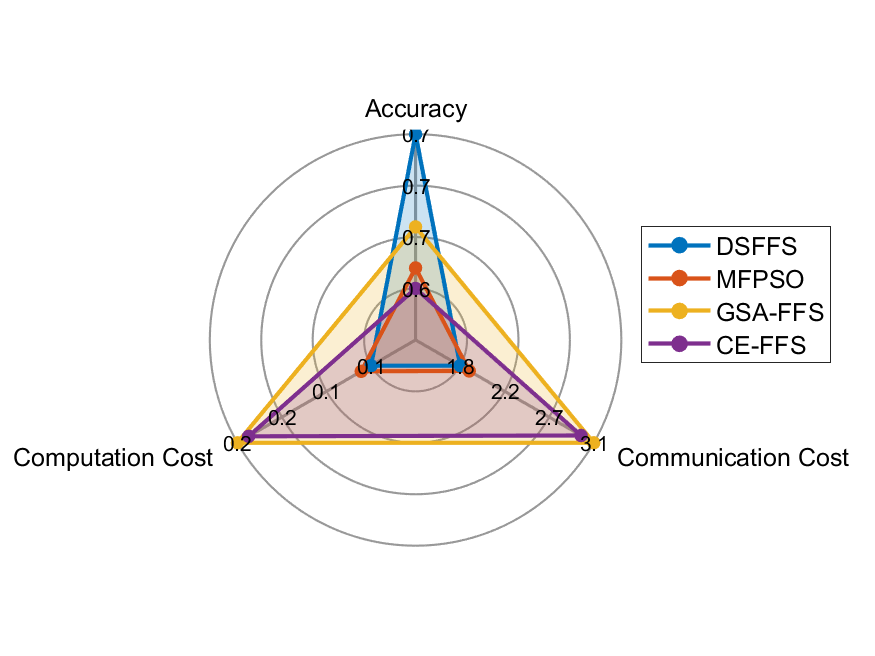}
      \vspace{-10mm}
      \caption{$Isolet$}
    \end{subfigure}&
    \begin{subfigure}{0.31\textwidth}
      \includegraphics[width=\textwidth]{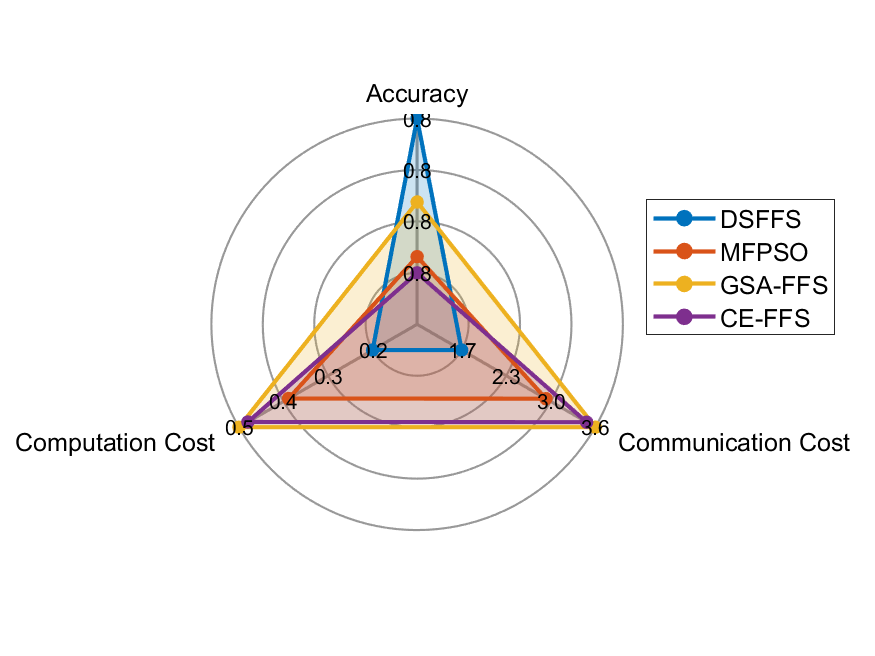}
      \vspace{-13mm}
      \caption{$MNIST$}
    \end{subfigure}&
    \begin{subfigure}{0.31\textwidth}
      \includegraphics[width=\textwidth]{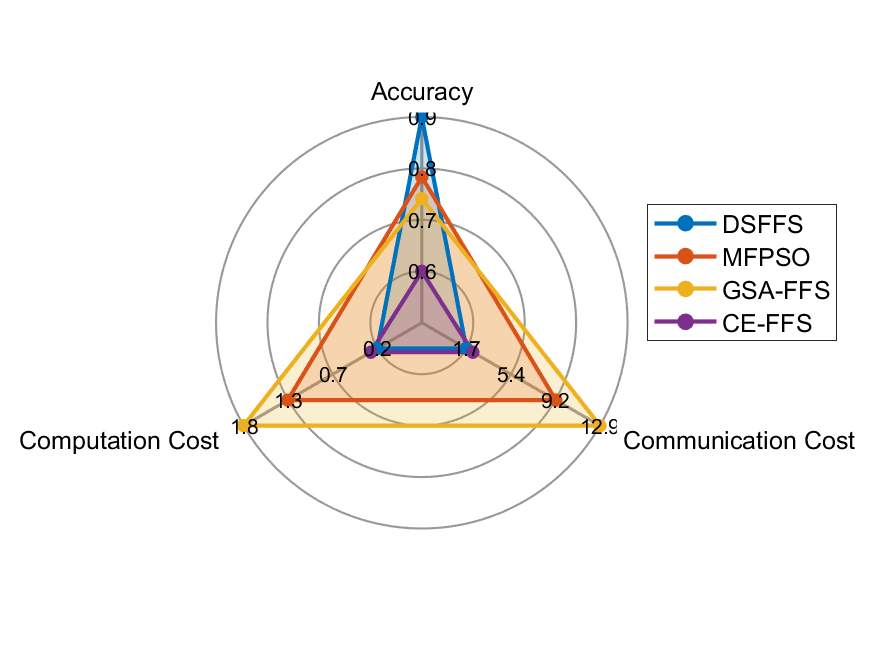}
      \vspace{-13mm}
      \caption{$PCMAC$}
    \end{subfigure}\\
    \begin{subfigure}{0.31\textwidth}
      \includegraphics[width=\textwidth]{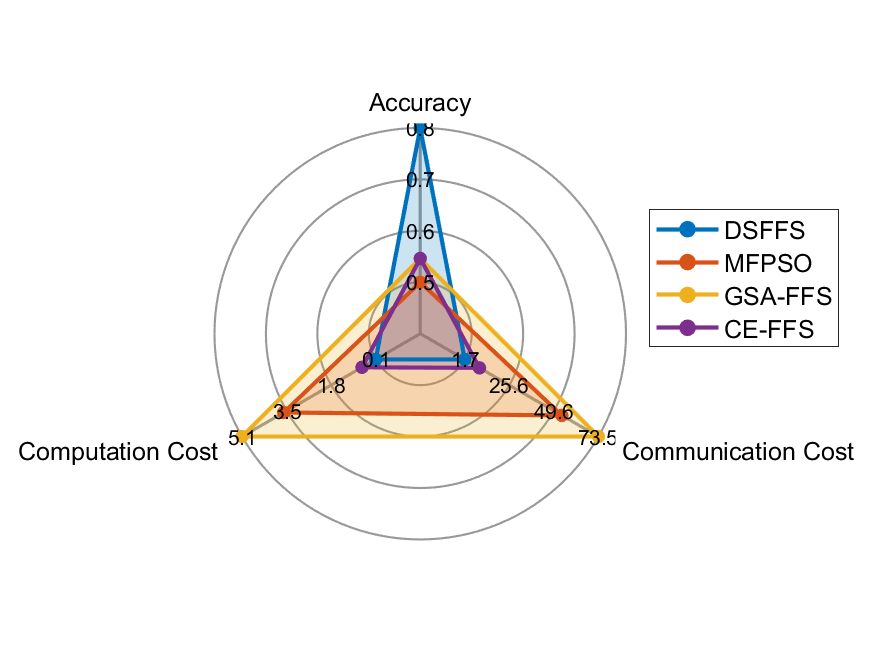}
      \vspace{-13mm}
      \caption{$SMK-CAN-187$}
    \end{subfigure}&
    \begin{subfigure}{0.31\textwidth}
      \includegraphics[width=\textwidth]{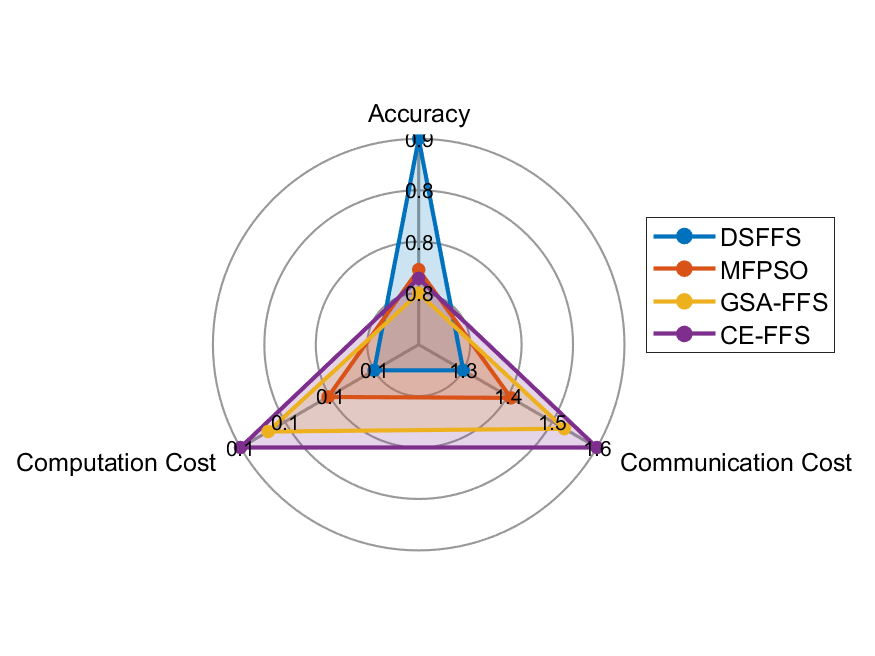}
      \vspace{-13mm}
      \caption{$USPS$}
    \end{subfigure}&
    \begin{subfigure}{0.31\textwidth}
      \includegraphics[width=\textwidth]{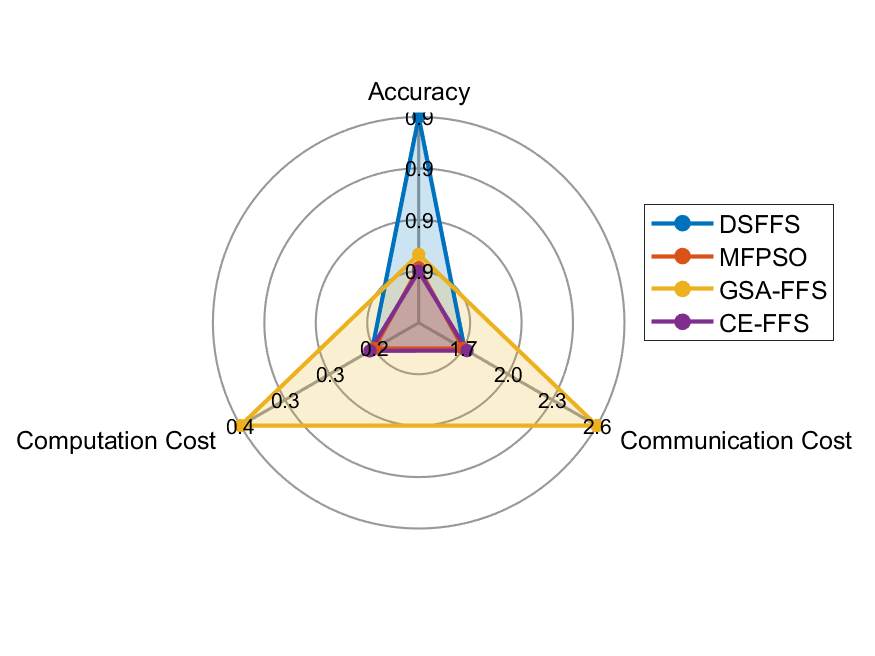}
      \vspace{-13mm}
      \caption{$HAR$}
    \end{subfigure}\\
    
    \end{tabular}
  \vspace{-3mm}
  \caption{Comparison of various evaluation metrics to assess the trade-off across all datasets}\label{fig:animals}
  \label{fig7}
\end{figure*}

\subsection{Results and Analysis}
\textbf{Classification Accuracy:} To evaluate the proposed method, we select different feature set sizes \(K \in \{50, 100, 150, 200\}\) and compare the performance of four sparse FL models: RandomMask, PruneFL \cite{jiang2022model}, FedDST, and FedDST+FedProx \cite{bibikar2022federated}. The results show that our method achieves the best performance when combined with FedDST and FedDST+FedProx, particularly at \(K=150\). When integrated with FedDST, the proposed DSFFS method outperforms other sparse FL models in 22 out of 36 cases across nine datasets and four feature set sizes. Additionally, when combined with FedDST+FedProx, it achieves superior performance in 8 out of 36 cases. Notably, with \(K=150\), our method combined with FedDST outperforms all other FL models in 8 out of 9 datasets. Due to space limitations, these results are not included here.

In Table \ref{table4}, we compare the performance of the proposed method with \(K=150\) features against three supervised FFS methods for horizontal FL: MFPSO, GSA-FFS, and CE-FFS. MFPSO and GSA-FFS are wrapper-based methods, while CE-FFS is a filter-based approach. The results show that our embedded method selects a more informative feature set, allowing the FL models, FedDST and FedDST+FedProx, to achieve higher performance with fewer selected features. For instance, on the \say{\(\texttt{PCMAC}\)} dataset, our method with 150 features attains an accuracy of 0.8688, whereas MFPSO, the second-best method, requires 1163 features to reach an accuracy of only 0.7480. Moreover, on the \say{\(\texttt{COIL-20}\)} dataset, the proposed method with 150 features achieves an accuracy of 0.8298, while GSA-FFS, using 505 features, achieves 0.7534.

\textbf{Computation and Communication Costs:} Table \ref{table4} presents the cumulative number of FLOPs per client across nine datasets for the proposed method and three comparative methods. The results show that the proposed approach significantly reduces computational costs in the FL process by maintaining sparse networks in both the input and hidden layers, thereby lowering local computational workloads.

Figures \ref{fig3} -- \ref{fig5} illustrate the performance of the sparse FL model (FedDST) in relation to cumulative upload costs between clients and the server, using DSFFS and three other federated feature selection methods on three datasets: \say{\(\texttt{MNIST}\)}, \say{\(\texttt{Fashion-MNIST}\)}, and \say{\(\texttt{COIL-20}\)}. The results demonstrate that the proposed method achieves superior performance with lower communication overhead. For instance, in Figure \ref{fig3}, on the \say{\(\texttt{MNIST}\)} dataset, the proposed method reaches an accuracy of 0.70 with an upload cost of just 0.01 GiB. In contrast, other methods—MFPSO, GSA-FFS, and CE-FFS—achieve lower accuracies of 0.50, 0.23, and 0.20, respectively, at the same upload cost. Similarly, as shown in Figures \ref{fig4} and \ref{fig5}, DSFFS outperforms the alternatives by achieving higher accuracy with lower upload costs.

\textbf{Overall Performance:} To comprehensively evaluate each method’s performance across multiple metrics—classification accuracy (\(\uparrow\)), upload cost (\(\downarrow\)), and computational complexity in terms of FLOPs (\(\downarrow\))—we assessed their effectiveness on all datasets. Figure \ref{fig7} summarizes the performance of each method across different datasets. While GSA-FFS generally achieves higher classification accuracy than MFPSO in most cases, MFPSO benefits from lower computation and communication costs. CE-FFS improves efficiency in both computation and communication but sacrifices classification accuracy. In contrast, the proposed method, DSFFS, consistently achieves the highest classification accuracy while maintaining the lowest computation and communication costs across most datasets.

The results demonstrate that DSFFS provides a good trade-off across all three evaluation metrics compared to other baselines. Specifically, DSFFS outperforms GSA-FFS in classification accuracy by 3.74\(\%\) on \say{\(\texttt{MNIST}\)} and 4.88\(\%\) on \say{\(\texttt{Fashion-MNIST}\)}, both of which contain a large number of samples. Furthermore, while some baseline methods perform well in specific domains, they may struggle in others. For instance, CE-FFS excels in biological datasets, whereas DSFFS maintains robust and consistent performance across datasets with varying characteristics. Therefore, in applications with diverse device heterogeneity, DSFFS can offer a well-balanced trade-off between model performance, computation, and communication costs while mitigating the impact of resource-limited devices by selecting informative features with an appropriate level of sparsity.

\section{Conclusion and Future Works}
In this paper, we introduce DSFFS, a novel and efficient embedded-based FFS method for horizontal FL. DSFFS dynamically updates input-layer neurons and connections, as well as hidden-layer connections, during training to identify informative features. Extensive experiments were conducted on nine diverse datasets with four different feature set sizes. Among 36 comparisons across four sparse FL methods, DSFFS combined with FedDST achieved the highest classification accuracy in 22 cases. Additionally, compared to three other supervised FFS methods, DSFFS with FedDST (\(K=150\)) outperformed them in 8 out of 9 datasets. Results further show that DSFFS significantly reduces communication and computation costs—by 76\(\%\) and 78\(\%\), respectively—on the non-iid Isolet dataset compared to GSA-FFS. Overall, DSFFS provides an effective trade-off between model accuracy, communication, and computation costs, particularly at high sparsity levels (e.g., 0.8, as used in this study). Future research will focus on extending DSFFS to sparse semi-supervised FL in non-iid data distribution settings.

\vspace{-2mm}
\small
\bibliographystyle{IEEEtranN}
\bibliography{References}
\end{document}